\documentclass[11pt,a4paper]{article}

\usepackage[utf8]{inputenc}
\usepackage[T1]{fontenc}
\usepackage[english]{babel}
\usepackage{lmodern}   

\usepackage[margin=1in]{geometry}
\usepackage{amsmath,amssymb,amsthm}
\usepackage{graphicx}
\usepackage{booktabs}
\usepackage{microtype}
\usepackage{xcolor}
\usepackage[colorlinks=true,linkcolor=blue,citecolor=blue,urlcolor=blue]{hyperref}
\usepackage{cleveref}
\usepackage[numbers,sort&compress,square]{natbib}
\usepackage{orcidlink}

\newcommand{\IM}{I\&M}

\title{From Black-Box Confidence to Measurable Trust in Clinical AI:\\
A Framework for Evidence, Supervision, and Staged Autonomy%
\thanks{This work has been submitted to the IEEE for possible publication.
Copyright may be transferred without notice, after which this version may no
longer be accessible.}}

\author{%
  Serhii Zabolotnii\,\orcidlink{0000-0003-0242-2234}\thanks{Corresponding author:
  \href{mailto:zabolotnii.serhii@csbc.edu.ua}{zabolotnii.serhii@csbc.edu.ua}}\\
  \small Cherkasy State Business College, Cherkasy, Ukraine\\
  \small healthPrecision, New York, NY, USA
  \and
  Viktoriia Holinko\,\orcidlink{0009-0001-1882-7503}\\
  \small healthPrecision, New York, NY, USA
  \and
  Olha Antonenko\,\orcidlink{0009-0009-5087-5394}\\
  \small healthPrecision, New York, NY, USA
}

\date{}   

\begin{document}
\maketitle

\begin{abstract}
\noindent
Trust in clinical artificial intelligence (AI) cannot be reduced to model
accuracy, fluency of generation, or overall positive user impression. In
medicine, trust must be engineered as a measurable system property grounded
in evidence, supervision, and operational boundaries of AI autonomy. This
article proposes a practical framework for trustworthy clinical AI built
around three principles: evidence, supervision, and staged autonomy. Rather
than replacing deterministic clinical logic wholesale with end-to-end
black-box models, the proposed approach combines a deterministic core, a
patient-specific AI assistant for contextual validation, a multi-tier model
escalation mechanism, and a human supervision layer for verification,
escalation, and risk control. We demonstrate that trust also depends on
selective verification of clinically critical findings, bounded clinical
context, disciplined prompt architecture, and careful evaluation on realistic
cases. Classifier-driven modular prompting is examined as an incremental path
to scaling clinical depth without sacrificing prompt performance and without
waiting for complete rule-based coverage. To operationalize trust, a set of
trust metrics is proposed, built on metrological principles --- measurement
uncertainty, calibration, traceability --- enabling quantitative rather than
subjective assessment of each architectural layer. In this perspective,
trustworthy clinical AI emerges not as a property of an individual model, but
as an architectural outcome of a system into which evidence trails, human
oversight, tiered escalation, and graduated action rights are embedded from
the outset.

\vspace{0.5em}
\end{abstract}

\section{Why trust has become the central problem of clinical AI}
\label{sec:intro}

Clinical artificial intelligence has rapidly moved beyond experimental
status. Today, systems based on large language models (LLMs) and other
generative approaches are increasingly entering medical documentation,
clinical message analysis, decision support, electronic health record (EHR)
management, and various forms of physician--patient communication. Yet a
central question grows alongside this expansion: not merely \emph{what} AI
can do, but \emph{when} and \emph{under what conditions} it can be trusted.

This question is especially acute in healthcare. Unlike many other domains,
the clinical context combines high error costs, time-sensitive decisions,
asymmetric risks, legal accountability, and the impossibility of fully
delegating professional judgment to a machine. Under such conditions, it is
insufficient for a model to produce plausible or even frequently helpful
responses. A clinically impressive output is not yet a trustworthy system.
From the perspective of instrumentation and measurement (\IM{}), this is
fundamentally a metrological problem of AI: how to calibrate, verify, and
determine measurement uncertainty for an intelligent system operating in a
clinical environment~\citep{who2021ethics,fda_aiml}. The International
Vocabulary of Metrology (VIM) defines \emph{measurand}, \emph{measurement
uncertainty}, and \emph{metrological traceability} as fundamental
characteristics of any measuring instrument. These concepts apply directly
to clinical AI: a system that ``measures'' the patient's state and generates
clinical recommendations. The GUM (Guide to the Expression of Uncertainty
in Measurement) provides a framework for the propagation of uncertainty
through a multi-stage measurement chain that maps naturally onto the
proposed multilayer architecture. In traditional metrology, no instrument
is introduced into clinical practice without calibration and confirmation
of fitness for purpose. An equivalent standard must apply to clinical AI.

\section{Current state of LLM use in medicine}
\label{sec:llm-state}

Contemporary literature shows that LLMs have already secured a prominent
place in clinical AI. They are entering most rapidly into segments where
value comes from working with language and large volumes of unstructured
data: clinical documentation, note summarization, message draft preparation,
patient education, information extraction from text, registry construction,
phenotyping, secondary review of records, and various forms of workflow
assistance. Review publications~\citep{bragazzi2024conceptual,santos2026cv}
indicate that interest in generative AI in medicine is no longer
hypothetical: these are not future possibilities, but systems already being
tested and partially deployed in practice.

At the same time, these same sources~\citep{bragazzi2024conceptual,rodrigues2025slr,santos2026cv}
reveal an important limitation: today LLMs perform significantly better as
augmentative tools than as autonomous clinical decision-makers. They can be
useful for draft generation, summarization, extraction support, triage
assistance, or explaining information, but remain insufficiently reliable
for independently executing high-stakes clinical decisions without serious
safeguards. This is especially true for personalized advice, deterministic
risk calculations, long reasoning chains, and situations where linguistic
plausibility is insufficient and auditability, stability, and accountability
are required.

\section{What the recent literature shows about key risks}
\label{sec:risks}

Recent literature is increasingly documenting that the central problem lies
not only in isolated hallucinations, but in a broader trust gap between the
impressive \emph{in silico} behavior of systems and their actual clinical
suitability. Consensus documents~\citep{futureai2025} emphasize that medical
AI requires more than simple accuracy: it needs traceability, robustness,
explainability, usability, and other systemic properties without which an
algorithm cannot be considered legitimately integrated into the clinical
process.

A distinct line of research highlights the risk of \emph{overtrust}. Work
in~\citep{shekar2025overtrust} shows that both patients and clinicians tend
to overestimate confident, well-formulated LLM outputs, even when these are
insufficiently accurate or safe. This means that the natural linguistic
persuasiveness of a model itself becomes a risk factor. Another important
body of literature concerns the \emph{accountability paradox}. Studies on
clinician-in-the-loop report generation and AI-assisted
documentation~\citep{zhang2026hitl} show that even when AI generates quality
text, the physician still must spend considerable time on full verification
due to inevitable professional and legal accountability. Therefore, high
generation quality alone does not eliminate review burden.

An additional important finding comes from work on tiered oversight and
hierarchical safety architectures~\citep{kim2025tao,li2026triad}. These
demonstrate that for complex clinical environments, the promising approaches
are not monolithic AI systems, but multilevel control models in which
different agents or different model classes perform different functions:
routing, screening, adjudication, escalation. This approach aligns well with
broader governance literature~\citep{li2026triad} and with practical
deployment logic in medical AI.

\section{Why this is insufficient without a trust architecture}
\label{sec:trust-architecture}

This is where the critical distinction between \emph{black-box confidence}
and \emph{measurable trust} emerges. The former is based on external
impression: the model speaks confidently, sounds professional, demonstrates
strong test results, creates a sense of competence. The latter is based on
architecture: can the recommendation logic be traced? Is the context
correctly selected? Is there an escalation mechanism? Are the limits of
permissible action defined? Is human control preserved? Does the system
withstand real workloads?

In medical AI this is not abstract philosophy but a direct condition of
safety. A system may demonstrate acceptable average quality while being
dangerous if its most critical errors occur in high-risk cases, if it
produces an excessive number of false positives, overloads human review,
or creates an illusion of autonomy where clinical intervention is actually
required.

\begin{figure}[t]
  \centering
  \includegraphics[width=0.85\linewidth]{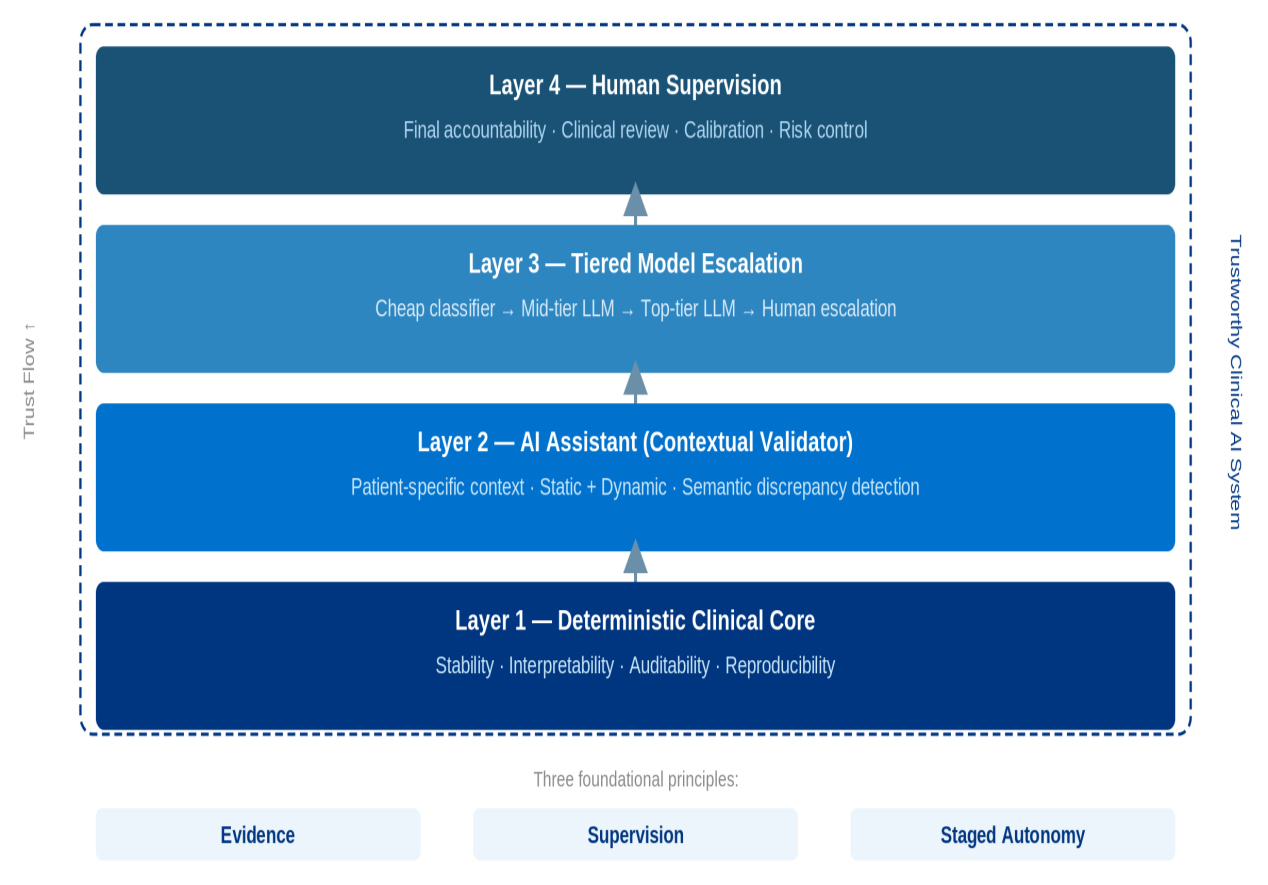}
  \caption{Layered architecture of trustworthy clinical AI.}
  \label{fig:architecture}
\end{figure}

This article argues that trustworthy clinical AI must be understood not as
a property of a model but as an architecturally engineered and operationally
measurable system property. The proposed framework rests on three principles:
\emph{evidence}, \emph{supervision}, \emph{staged
autonomy}~\citep{futureai2025,who2021ethics,fda_aiml}. In this perspective,
trust arises when evidence trails, human oversight, tiered escalation,
bounded context, and policy-based action rights are embedded in the system
design from the outset. The overall logic of this architecture is summarized
in \cref{fig:architecture}.

\section{From model performance to measurable trust}
\label{sec:measurable-trust}

One of the most common errors in discussions about medical AI is reducing
trust to model quality indicators. Accuracy, F1-score, AUC, human preference
ratings, or general impression of a response matter, but they do not capture
what actually determines safe clinical use.

In a real care environment, trust depends on at least several dimensions.
The first is \textbf{analytical validity}. The system must stably perform
its task under realistic conditions. The second is \textbf{contextual
appropriateness}. Even a high-quality model can generate an unacceptable
recommendation if it relies on the wrong time horizon, the wrong clinical
background, or incomplete patient-specific information. The third is
\textbf{evidence traceability}. For clinically important decisions, having
the answer is insufficient; it must be possible to understand what data,
features, and mechanisms produced it. The fourth is \textbf{reproducibility
and stability}. The system must behave predictably on similar cases and not
undermine trust through chaotic variability. The fifth is \textbf{workflow
fit}. Even a powerful AI is of little use if its outputs are inconvenient
for the physician, arrive at the wrong moment, or require disproportionate
additional work. The sixth is a \textbf{safe operational zone}. It is
essential to clearly define what the system is and is not permitted to do.

Each of these dimensions has a precise metrological analog. Analytical
validity corresponds to the accuracy and precision of a measuring instrument.
Contextual appropriateness corresponds to the influence of environmental
conditions on measurement. Evidence traceability corresponds to metrological
traceability to reference standards. Reproducibility and stability correspond
to repeatability and reproducibility in the GUM sense. Workflow fit
corresponds to the usability of a measurement system in an operational
environment. Safe operational zone corresponds to the measurement range and
operational limits of an instrument. This correspondence is not a superficial
analogy: it enables borrowing from the \IM{} discipline the ready-made tools
for specification, verification, and quality control of clinical AI.

This approach changes the very formulation of the problem. Instead of asking
``How intelligent is the model?'', one must pose more complex and correct
questions: Can the output be verified? Is the context correctly selected?
Is there an explicit escalation path? Do some errors carry disproportionately
greater weight? Who reviews suspicious results and when? Is the system's
behavior aligned with clinical workflow?

Thus, trustworthy clinical AI must be \emph{engineered}, not
\emph{assumed}~\citep{futureai2025,li2026triad}. Trust is not the natural
consequence of a larger model or a more elegant interface. It arises as the
result of deliberate design choices that determine the interaction among
data, reasoning, verification, escalation, action, and human accountability.

\section{Practical architecture: deterministic core, AI assistant, model escalation, and human supervision}
\label{sec:architecture}

One practical way to operationalize trust is to build a multilayer
architecture in which different components perform different functions and
have different responsibility boundaries.

\subsection{Deterministic clinical core as a trust anchor}
The foundation of such a system must remain the \textbf{deterministic
clinical core} --- a rule-based or hybrid core that ensures stability,
interpretability, and controllability of clinical logic. In the era of
foundation models this may sometimes seem ``old-fashioned,'' but for
healthcare precisely such logic is often critically important. It establishes
a structured backbone onto which more flexible AI layers can be grafted
without losing basic predictability.

The value of such a core lies not in maximum flexibility, but in something
else: it provides clear rules, control over updates, behavioral
reproducibility, and better auditability~\citep{futureai2025,bragazzi2024conceptual}.
In the clinical context this is not a limitation but a support.

\subsection{AI assistant as contextual validator}
The second layer is the \textbf{AI assistant}. It is best understood not as
an autonomous decision-maker, but as a contextual validator, semantic
reviewer, discrepancy detector, or correction layer. Its strength lies not
in fully replacing deterministic clinical logic, but in reinforcing it where
weakly formalized, communicatively rich, or semantically ambiguous
situations arise.

This is where LLM-like systems can deliver real value: analyzing patient
messages, noticing contextual inconsistencies, correlating recommendations
with recent events, synthesizing dispersed signals, or helping prepare
provider-facing communication.

For this to work safely, the AI assistant must be \textbf{patient-specific}.
This requires combining at least two types of context: \emph{static context}
--- diagnoses, medications, allergies, medical history, stable clinical
characteristics; \emph{dynamic context} --- recent events, current changes,
recent messages, alerts, near-term workflow signals.

This distinction matters because not all clinical information ages at the
same rate. Without this discipline, the system either drowns in stale context
or loses critically important recent signals.

\subsection{Tiered model escalation: not all LLMs should do the same thing}
\begin{figure}[t]
  \centering
  \includegraphics[width=0.85\linewidth]{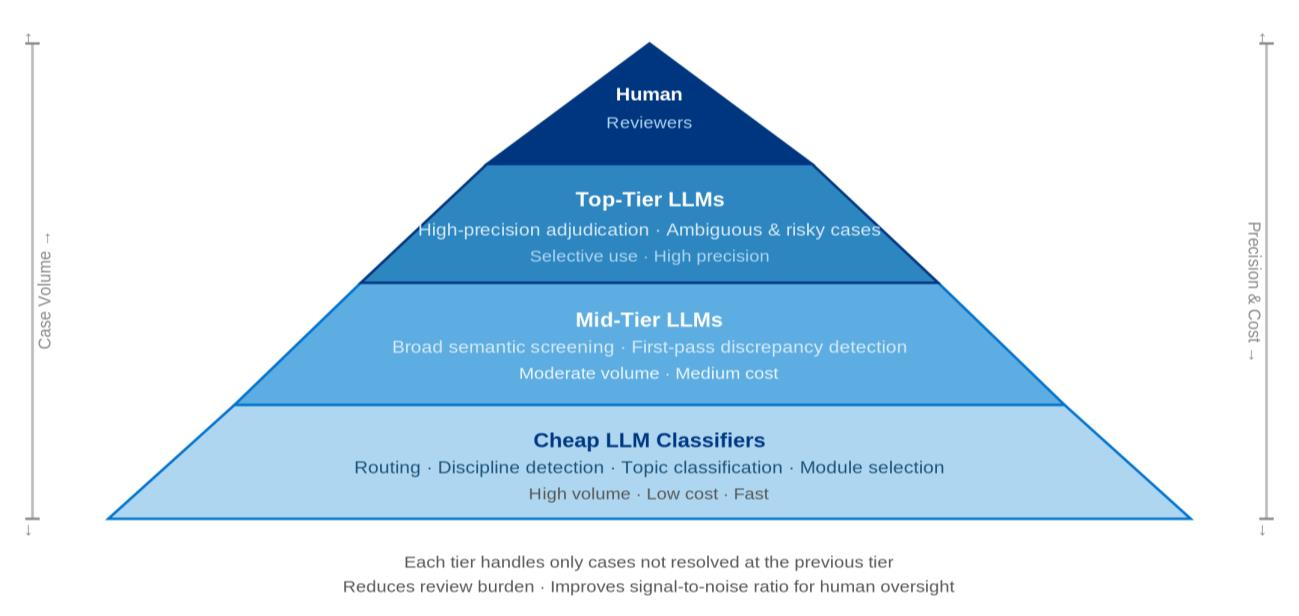}
  \caption{Role-specialized model hierarchy for scalable trust.}
  \label{fig:hierarchy}
\end{figure}

Another important layer that fundamentally reinforces the trust architecture
is a \textbf{role-specialized tiered model hierarchy}. In real high-volume
clinical workflows, it is irrational to apply top-tier expensive models to
all documents, messages, and cases. This is not only financially demanding
but often unnecessary. At the same time, a direct jump from a mid-class LLM
to human review is equally problematic: if the system produces many false
positives, it quickly overloads human reviewers and degrades the
signal-to-noise ratio of supervision. This multilevel escalation logic is
separately visualized in \cref{fig:hierarchy}.

It is therefore advisable to build tiered escalation~\citep{kim2025tao}:
\emph{cheap LLM classifiers} --- for routing, discipline detection, module
selection, topic classification; \emph{mid-tier LLMs} --- for broad semantic
screening and first-pass discrepancy detection; \emph{top-tier LLMs} --- for
high-precision adjudication on selected ambiguous or risky cases only;
\emph{human reviewers} --- only for genuinely high-risk, unresolved, or
high-stakes situations.

This architecture means that trustworthy clinical AI must have not only a
\emph{Human-in-the-Loop}, but also a \emph{model-tier-in-the-loop}. Cheaper
models can handle routing and initial classification. Mid-class models can
process the broad stream and perform broad review. Top-tier models can
function as an expert adjudication layer before human intervention. Only
after this do cases reach the human reviewer.

This matters for three reasons. First, this approach reduces cost. Second,
it decreases review burden. Third, it makes supervision more precise: humans
receive less noise and more genuinely critical cases.

\subsection{Human supervision as a constructive architectural layer}
The final and fundamentally necessary layer is \textbf{human supervision}.
In clinical AI, the human must not appear only as an emergency safeguard.
Human-in-the-loop is a constructive, not cosmetic, system layer. It fulfills
the function of final accountability, clinical review, correction of
systemic errors, and calibration of where AI authority
ends~\citep{zhang2026hitl,li2026triad}.

Thus, in this multilayer architecture: the deterministic core provides
stability; the AI assistant adds contextual sensitivity; tiered model
escalation manages scaling, cost, and noise; human supervision preserves
accountability and bounded control.

This is how trust transforms from an abstract concept into a practical
system architecture.

\section{Trust as graduated authority}
\label{sec:graduated-authority}

If trust is truly a measurable system property, it must influence not only
model evaluation but also the boundaries of permissible action. This is the
idea of \textbf{graduated authority} or \textbf{staged autonomy}.

In many AI systems, autonomy is viewed as the natural next step: once a
model demonstrates acceptable quality, it is expected to be permitted to do
more. In healthcare, this thinking is dangerous. Autonomy must not be
granted in advance. It must be earned through accumulated evidence,
stability, risk control, and real review outcomes~\citep{santos2026cv,li2026triad}.

\paragraph{Level 1 --- Monitoring and Alerting.}
At the first level, the system does not act directly but observes. The AI
assistant analyzes the outputs of the deterministic core, compares them with
patient context, and identifies discrepancies or suspicious cases. Its
primary task here is to \emph{notice}, not to intervene.

\paragraph{Level 2 --- Validated Recommendation Support.}
At the second level, when sufficient confirmation of reliability exists, AI
can move to a more active but still controlled role. It can: propose
alternative formulations; highlight mismatches; prioritize escalation; help
structurally prepare provider-facing communication. The key constraint
remains that output is still within a controlled reviewable process.

\paragraph{Level 3 --- Limited Direct Operational Participation.}
Only after convincing confirmation of safe and stable behavior can limited
direct system participation in operational workflows be considered. And even
then, the discussion should not be about full autonomy, but about
\emph{constrained direct participation} in clearly defined use cases and
under continued supervision.

\begin{figure}[t]
  \centering
  \includegraphics[width=0.85\linewidth]{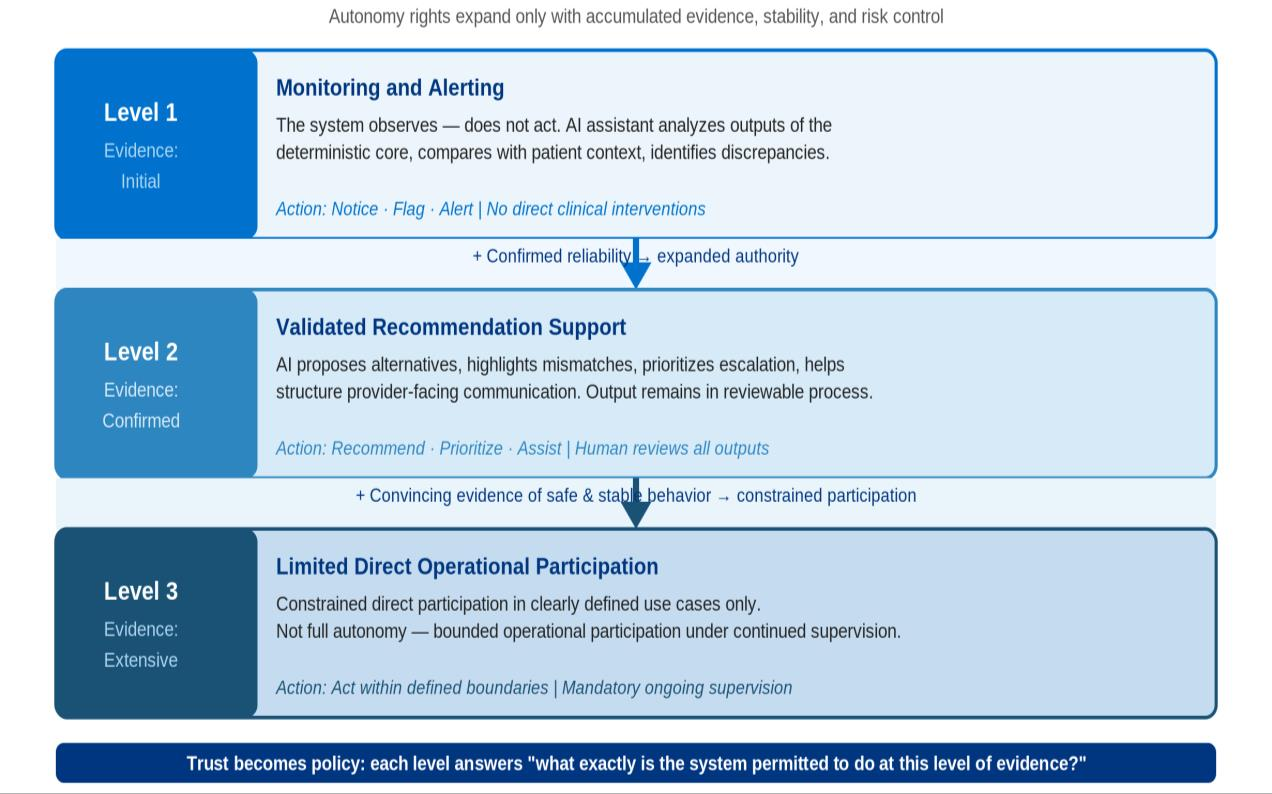}
  \caption{Graduated authority in clinical AI.}
  \label{fig:authority}
\end{figure}

This model matters because it transforms trust into policy. Trust ceases to
be a general rating and becomes an answer to a specific question: what
exactly is the system permitted to do \emph{at this level of evidence}? The
idea of stepwise expansion of action rights is directly illustrated in
\cref{fig:authority}.

\section{Selective verification and bounded clinical context}
\label{sec:bounded-context}

Another central thesis of the article is that \textbf{not all errors are
equally important}. In clinical AI, it is irrational to apply uniformly
intensive verification effort to all outputs. Some errors are regrettable
but minor; others can have disproportionately high clinical consequences.

This is why \textbf{selective verification of clinically critical
findings}~\citep{zhang2026hitl} is necessary. Instead of total rechecking,
risk-aware design makes increasing sense, in which additional verification
is triggered where: clinical risk of error is high; ordinary rule-based logic
has weak points; additional semantic review is operationally justified.

This same bounded principle applies to context. Clinical AI does not simply
need \emph{more} data. It needs \emph{correctly selected clinical context}.
Relevant context is part of safety. A poorly chosen background can destroy
the quality of even a strong model --- through distraction, stale detail
overload, or loss of focus on what truly matters.

This is why patient summaries or clinical backgrounds must be built as a
\emph{bounded clinical state representation}, not as a cumulative archive.
Such a summary must support: replacement of outdated data; pruning and
deduplication; different time horizons for different types of information;
retention of only what is clinically relevant now.

The same applies to chat history. A wider context window is not always
better. In many cases, a shorter task-bounded segment of history is more
useful than a long transcript that merely inflates prompt size, latency, and
ambiguity.

\begin{figure}[t]
  \centering
  \includegraphics[width=0.85\linewidth]{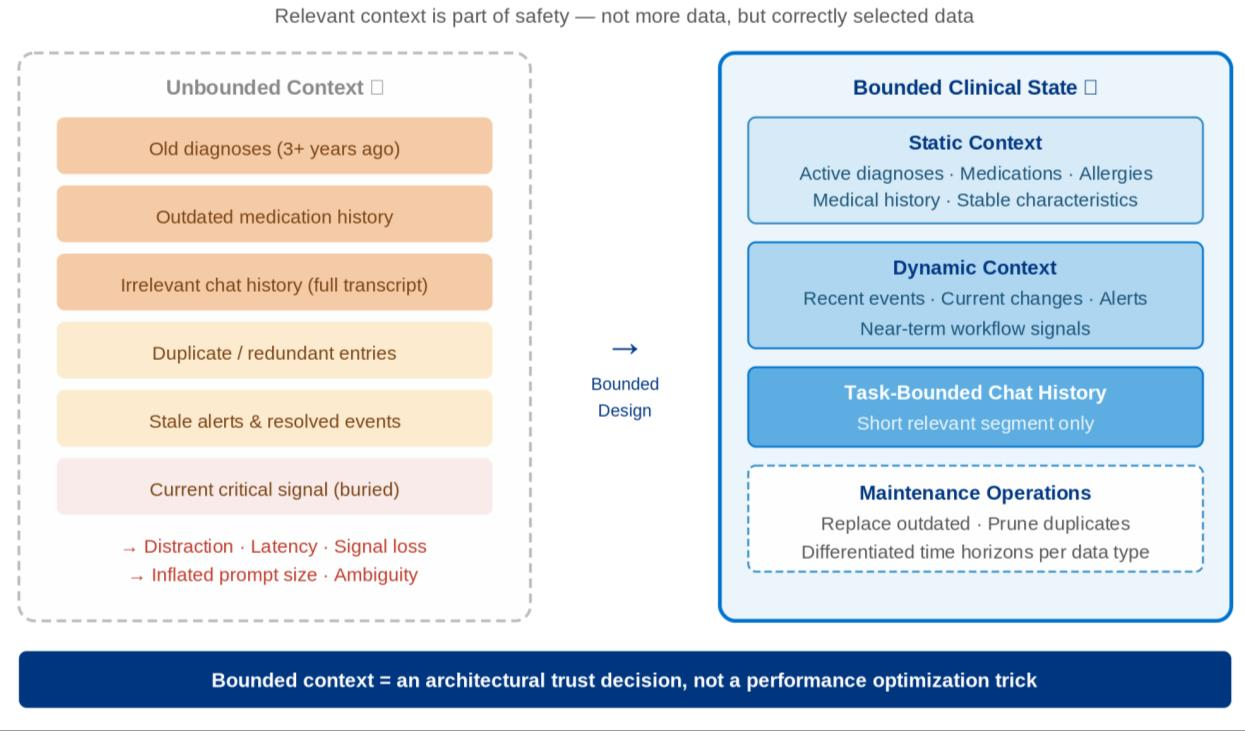}
  \caption{Bounded clinical background design.}
  \label{fig:bounded}
\end{figure}

Thus, bounded context is not an optimization trick, but a part of trust
architecture (see \cref{fig:bounded}).

\section{Prompt architecture as part of trust architecture}
\label{sec:prompt-architecture}

In clinical AI, prompting cannot be regarded as a superficial layer.
\textbf{Prompt architecture is part of the system architecture}, and
therefore part of safety.

One of the central problems here is \emph{prompt sprawl}. When increasingly
more rules, examples, scoring systems, exceptions, and domain-specific
instructions are gradually added to a single prompt, it begins to lose
clarity, stability, and manageability. Eventually an architectural ceiling
is reached: the model holds instructions less effectively, developers find
it harder to maintain the logic, and alignment with backend behavior becomes
increasingly unreliable.

This is why \textbf{modular prompt decomposition} is often a better path
than monolithic instruction blocks. Separate prompts or instruction modules
for different target contexts reduce the cognitive load on the model,
improve consistency, and facilitate auditability. Structured output schemas
further reinforce trust by making results machine-readable and suitable for
downstream processing.

Particularly promising is \textbf{classifier-driven modular architecture}.
Instead of encoding all discipline-specific knowledge in a single prompt, a
separate classifier first identifies the relevant clinical topic or workflow
context, then only the appropriate instruction modules are added at runtime.
This allows scaling clinical depth \emph{horizontally}, through modules,
rather than \emph{vertically}, through the endless growth of a single
prompt.

This approach is important not only from a performance perspective. It also
creates a realistic path toward a \emph{module-by-module clinical safety
net}: new knowledge modules can be added incrementally without disrupting
the baseline stability of the system.

\section{Trust metrics: a metrological approach to measuring trust}
\label{sec:trust-metrics}

\begin{figure}[t]
  \centering
  \includegraphics[width=0.85\linewidth]{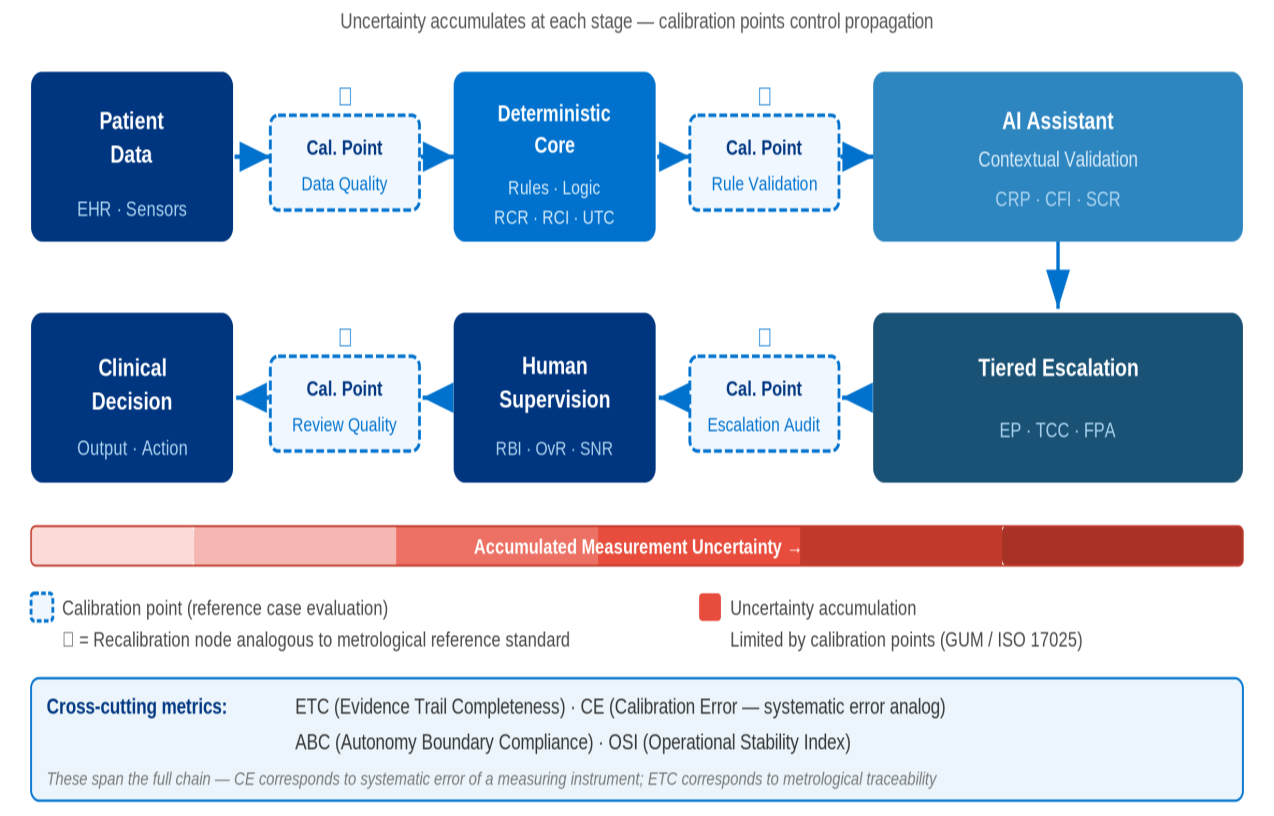}
  \caption{Metrological measurement chain for clinical AI trust.}
  \label{fig:chain}
\end{figure}

If trustworthy clinical AI is an architecturally engineered system property,
the practical question arises: how should this property be measured? Trust
that remains conceptual is not operational. The metrological approach
requires that each architectural layer have explicitly defined quality
indicators --- analogously to how each component of a measurement chain is
specified through its own metrological characteristics. The analogy between
a classical measurement chain and the clinical AI trust chain can be
illustrated schematically (see \cref{fig:chain}). At each stage of the chain
--- from patient data collection to clinical decision --- calibration points
and sources of uncertainty that accumulate along the chain are identified.

The following is a set of trust metrics structured according to four
architectural layers. These metrics constitute a conceptual framework for
further empirical validation --- they define \emph{what} should be measured
and serve as a starting point for specifying concrete measurement
methodologies and acceptable threshold values.

\paragraph{Layer 1 --- Deterministic Core.}
\textit{Rule Coverage Rate} (RCR) --- the proportion of clinical scenarios
covered by explicit rules; \textit{Rule Consistency Index} (RCI) --- stability
of rule outputs under system updates; \textit{Update Traceability Coefficient}
(UTC) --- the proportion of rule changes supported by documented clinical
rationale.

\paragraph{Layer 2 --- AI Assistant.}
\textit{Context Relevance Precision} (CRP) --- the proportion of context
elements genuinely relevant to the clinical question; \textit{Context
Freshness Index} (CFI) --- weighted timeliness of clinical data in the
prompt; \textit{Semantic Consistency Rate} (SCR) --- the proportion of
responses stable under input rephrasing.

\paragraph{Layer 3 --- Tiered Escalation.}
\textit{Escalation Precision} (EP) --- the proportion of escalations that
genuinely required a higher tier; \textit{Tier Cost Coefficiency} (TCC) ---
cost of processing one case at each tier; \textit{False Positive Attenuation}
(FPA) --- the reduction coefficient of false positives between adjacent
tiers.

\paragraph{Layer 4 --- Human Supervision.}
\textit{Review Burden Index} (RBI) --- average time per reviewed case;
\textit{Override Rate} (OvR) --- the proportion of AI outputs modified by a
human reviewer; \textit{Signal-to-Noise Ratio} (SNR) --- the ratio of
critical cases to total flow reaching the human layer.

Four cross-cutting metrics characterizing the system as a whole can be
separately identified: \textit{Evidence Trail Completeness} (ETC) --- the
proportion of outputs with a complete evidence tracing chain;
\textit{Calibration Error} (CE) --- deviation between the model's stated
confidence and its actual accuracy; \textit{Autonomy Boundary Compliance}
(ABC) --- the proportion of actions within defined autonomy rights;
\textit{Operational Stability Index} (OSI) --- variation of key metrics over
time as an indicator of long-term stability.

\begin{figure}[t]
  \centering
  \includegraphics[width=0.85\linewidth]{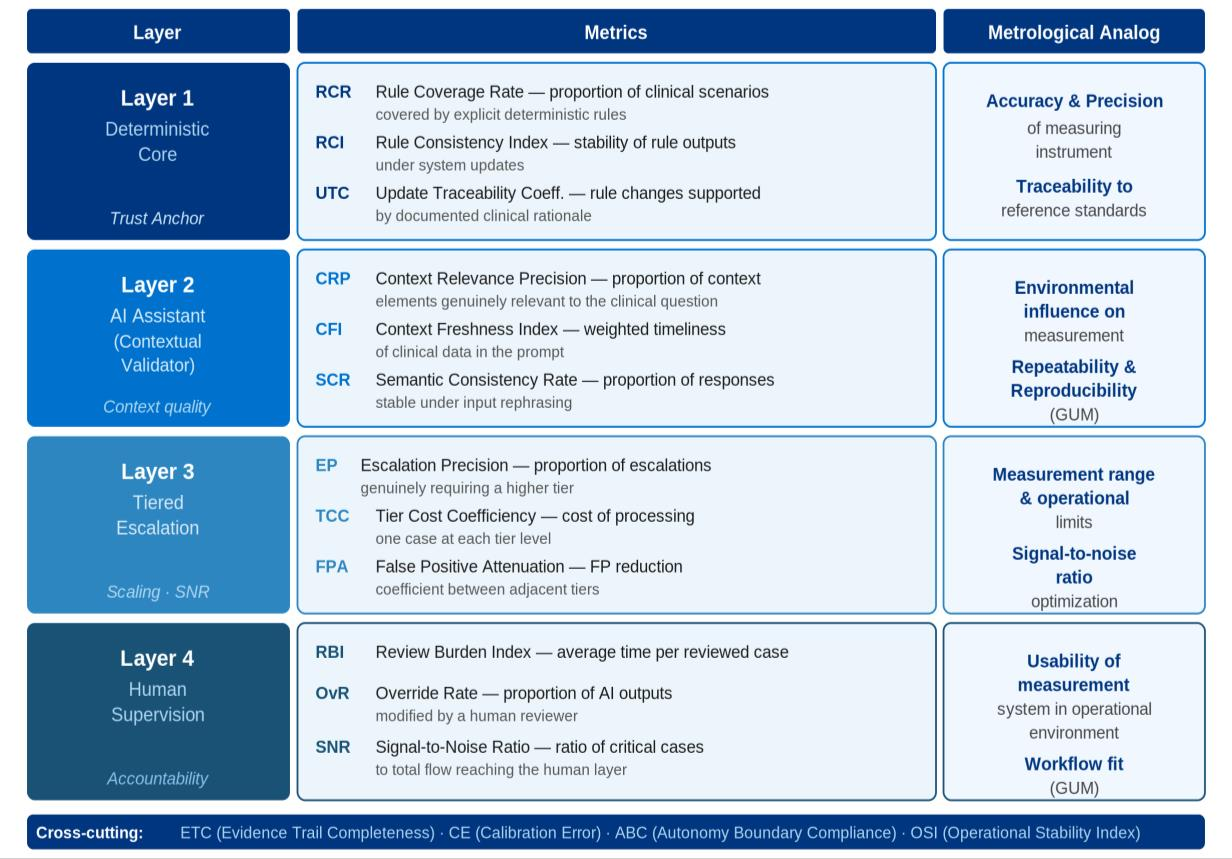}
  \caption{Proposed trust metrics framework mapped to architectural layers.}
  \label{fig:map}
\end{figure}

The distribution of all metrics across architectural layers is shown in
\cref{fig:map}. Calibration Error is a direct analog of the systematic error
of a measuring instrument and requires regular recalibration --- that is,
evaluation on new reference cases. Evidence Trail Completeness corresponds
to Metrological Traceability: the ability to trace any result to a
documented source. The Operational Stability Index corresponds to Stability
Testing and drift monitoring as adopted in metrological practice.

\section{Measured evaluation for measurable trust}
\label{sec:measured-evaluation}

If trust is to be measurable, evaluation must also be measured. In clinical
AI, relying solely on demo-driven impressions or a few successful outputs
is insufficient. Meaningful evaluation must include: model comparison on
identical cases; response consistency; time and cost characteristics;
medical expert review; realistic edge cases; production-derived scenarios;
explicit error analysis.

This matters not only for model selection, but for understanding how a
system behaves in the real world. Trustworthy clinical AI requires not only
strong runtime behavior, but a trustworthy testing architecture. Multi-agent
simulation, structured scenario validation, selective stress testing, and
review of real-world failures can be no less important than standard
benchmark evaluations.

In metrological terms, evaluation of clinical AI is \emph{calibration}:
verifying the system against reference cases analogously to verifying a
measuring instrument against a standard. Reference cases play the role of
calibration standards. Periodic re-evaluation corresponds to \emph{recalibration}
--- a mandatory procedure for any clinical measuring device. Monitoring
performance in production corresponds to \emph{stability testing} and
\emph{drift detection} as specified in GUM and ISO 17025. In this sense,
trust begins before deployment. It begins with how seriously a system
undergoes testing, calibration, and reality-based verification.

\section{Practical implications for teams building clinical AI}
\label{sec:implications}

The proposed framework has direct implications for developers, clinical
informatics teams, and healthcare organizations.

\emph{First}, it proposes changing the primary objective. The goal should
not be ``to give AI as much autonomy as quickly as possible.'' Instead, the
goal should be \textbf{bounded intelligence} that can be verified, explained,
constrained, and progressively scaled.

\emph{Second}, it demonstrates that building a trust architecture is not
only about the model. It is also about: quality of context design; prompt
discipline; output structure; escalation routing; review burden; physician
workflow fit.

\emph{Third}, this framework underscores the importance of physician
autonomy. Clinicians are far more willing to trust systems that maintain
clear responsibility boundaries, reduce workload, and support rather than
displace judgment~\citep{shekar2025overtrust}. In medicine, trust is
inseparable from control.

\emph{Fourth}, the proposed approach provides a practical way to resist
premature autonomy. Instead of asking whether a system is sufficiently
``intelligent,'' it is better to ask: What evidence is sufficient for the
next level of action? What supervision layer is still needed? Which errors
are disproportionately dangerous? Is the system overloading human review
with unnecessary noise? Are roles distributed across model tiers in an
economically and clinically sound manner?

\section{Conclusion}
\label{sec:conclusion}

Trustworthy clinical AI should not be understood as a property of an
individual model. It is the outcome of an architecture in which evidence
trails, human supervision, bounded context, disciplined prompting, measured
evaluation, tiered escalation, and graduated action rights are deliberately
embedded into the system.

In such a system, deterministic logic and generative AI are not adversaries.
They fulfill different but complementary roles. The deterministic core
stabilizes and constrains. The AI assistant adds contextual sensitivity.
Multilevel model escalation manages scaling, cost, and noise. Human
supervision ensures final accountability and preserves clinical control.

Thus, the path forward for clinical AI is not black-box confidence, but
measurable trust. From the \IM{} perspective, this means that clinical AI
must be treated as a measurement system for which the measurand, operational
limits, measurement uncertainty, calibration procedures, and fitness criteria
are defined~\citep{who2021ethics,fda_aiml}. The proposed trust metrics
framework --- from Rule Coverage Rate and Context Freshness Index to
Calibration Error and Evidence Trail Completeness --- is a first step toward
operational specification of these characteristics. It enables the
transition from subjective impressions of quality to measurable indicators
verified against reference cases. And it is precisely this shift --- from
impression to architecture, from smooth generation to evidence-based
controllability, from model performance to metrological traceability ---
that will determine which AI systems can be responsibly integrated into
real clinical practice.

\section*{Acknowledgment of AI use}
This article used Anthropic's Claude Sonnet 4.6 LLM in two capacities:
(1)~generating structural diagram visualizations (Figs.~1--6), and
(2)~editing and grammar enhancement of the manuscript text.

\bibliographystyle{unsrtnat}
\bibliography{references}

@article{futureai2025,
  author  = {{The FUTURE-AI Consortium}},
  title   = {{FUTURE-AI}: International consensus guideline for trustworthy and deployable artificial intelligence in healthcare},
  journal = {BMJ},
  year    = {2025},
  note    = {Consensus guideline}
}

@article{shekar2025overtrust,
  author  = {Shekar, Shruthi and Pataranutaporn, Pat and Sarabu, Chetanya and Cecchi, Guillermo A. and Maes, Pattie},
  title   = {People Overtrust {AI}-Generated Medical Advice despite Low Accuracy},
  journal = {NEJM AI},
  volume  = {2},
  number  = {6},
  year    = {2025},
  doi     = {10.1056/AIoa2300015}
}

@article{bragazzi2024conceptual,
  author  = {Bragazzi, Nicola Luigi and Garbarino, Sergio},
  title   = {Toward Clinical Generative {AI}: Conceptual Framework},
  journal = {JMIR AI},
  volume  = {3},
  pages   = {e55957},
  year    = {2024}
}

@article{kim2025tao,
  author       = {Kim, Youngjun and others},
  title        = {Tiered Agentic Oversight: A Hierarchical Multi-Agent System for Healthcare Safety},
  journal      = {arXiv preprint},
  year         = {2025},
  eprint       = {2506.12482},
  archivePrefix= {arXiv},
  primaryClass = {cs.AI}
}

@article{zhang2026hitl,
  author       = {Zhang, Xiaoyu and others},
  title        = {Human-in-the-Loop Interactive Report Generation for Chronic Disease Adherence},
  journal      = {arXiv preprint},
  year         = {2026},
  eprint       = {2601.06364},
  archivePrefix= {arXiv},
  primaryClass = {cs.CL}
}

@article{rodrigues2025slr,
  author  = {Rodrigues, Tiago and Teixeira Lopes, Carla},
  title   = {Harnessing Large Language Models for Clinical Information Extraction: A Systematic Literature Review},
  journal = {ACM Transactions on Computing for Healthcare},
  year    = {2025}
}

@article{santos2026cv,
  author  = {Ferreira Santos, Jo{\~a}o and Dores, Hugo},
  title   = {Large Language Models in Cardiovascular Prevention: A Narrative Review and Governance Framework},
  journal = {Diagnostics},
  volume  = {16},
  number  = {3},
  pages   = {390},
  year    = {2026},
  doi     = {10.3390/diagnostics16030390}
}

@article{li2026triad,
  author  = {Li, Jia and Zhou, Z. C. and Wang, Z. C. and Lv, H.},
  title   = {Prioritizing Human--{AI} Collaboration in Healthcare: The {TRIAD} Framework for Trustworthy Governance, Real-World, and Integrated Adaptive Deployment},
  journal = {Military Medical Research},
  volume  = {12},
  pages   = {97},
  year    = {2026},
  doi     = {10.1186/s40779-026-00684-w}
}

@techreport{who2021ethics,
  author      = {{World Health Organization}},
  title       = {Ethics and Governance of Artificial Intelligence for Health: {WHO} Guidance},
  institution = {WHO},
  address     = {Geneva, Switzerland},
  year        = {2021}
}

@misc{fda_aiml,
  author       = {{U.S. Food and Drug Administration}},
  title        = {Artificial Intelligence and Machine Learning ({AI}/{ML})-Enabled Medical Devices},
  howpublished = {\url{https://www.fda.gov/medical-devices/software-medical-device-samd/artificial-intelligence-and-machine-learning-aiml-enabled-medical-devices}},
  note         = {Accessed: 9 April 2026},
  year         = {2026}
}

\end{document}